\documentclass{bmvc2k}

\usepackage{multirow}
\usepackage{microtype}
\usepackage{booktabs}
\usepackage{tabularx}
\usepackage{soul}
\usepackage{pifont}
\usepackage{hyperref}
\hypersetup{
    colorlinks=true,
    linkcolor=blue,
    filecolor=magenta,      
    urlcolor=cyan,
}
 
\newcommand{\cmark}{\ding{51}}%
\newcommand{\xmark}{\ding{55}}%

\title{Simple vs complex temporal recurrences for video saliency prediction}

\addauthor{Panagiotis Linardos}{linardos.akis@gmail.com}{1}
\addauthor{Eva Mohedano}{eva.mohedano@insight-centre.org}{1}
\addauthor{Juan Jose Nieto}{juanjo.3ns@gmail.com}{1}
\addauthor{Noel E. O'Connor}{noel.oconnor@insight-centre.org}{1}
\addauthor{Xavier Giro-i-Nieto}{xavier.giro@upc.edu}{2}
\addauthor{Kevin McGuinness}{kevin.mcguinness@insight-centre.org}{1}

\addinstitution{
 Insight Center for Data Analytics\\
 Dublin City University\\
 Dublin, Ireland
}
\addinstitution{
 IDEAI-GPI\\
 Universitat Politecnica de Catalunya\\
 Barcelona, Catalonia/Spain
}

\runninghead{Linardos et al}{Temporal recurrences for video saliency prediction}

\def\etal{\emph{et al}\bmvaOneDot}

\begin{document}
\maketitle

\begin{abstract}
This paper investigates modifying an existing neural network architecture for static saliency prediction using two types of recurrences that integrate information from the temporal domain. The first modification is the addition of a ConvLSTM within the architecture, while the second is a conceptually simple exponential moving average of an internal convolutional state. We use weights pre-trained on the SALICON dataset and fine-tune our model on DHF1K. Our results show that both modifications achieve state-of-the-art results and produce similar saliency maps. Source code is available at \href{https://git.io/fjPiB}{https://git.io/fjPiB}.
\end{abstract}


\section{Introduction}
\label{sec:intro}

Visual saliency pertains to how an object or any piece of information may stand out from its surroundings. Detecting saliency is an integral part of how sentient organisms process information. We live in a world where the visual data we receive on a daily basis is immense and cluttered with noise; therefore, the brain has evolved in such a way that allows living organisms to focus their attention on the most relevant information, so as to function efficiently. Efforts in the computer vision community have been ongoing for many years to simulate this biological process artificially leading to the development of large-scale static gaze datasets, (e.g. SALICON~\cite{SALICON}) and, more recently, dynamic gaze datasets (e.g. DHF1K~\cite{Wang2018a}). Based on these datasets, model-driven approaches tackle the task of saliency prediction by estimating heatmaps of probabilities, where every probability corresponds to how likely it is that the corresponding pixel will attract human attention. Thanks to the availability of large-scale datasets, deep learning architectures have managed to significantly improve the accuracy achievable in this task (e.g.~\cite{Wang2018a,Pan2017,gorji2018going,SALICON,pan2016shallow}). 

Most scientific interest has so far been focused on image-based saliency models, with video saliency prediction gaining more traction in recent years with the introduction of large-scale video saliency datasets (\cite{Wang2018a, Hollywood_UCF}). When it comes to extracting visual information from the temporal domain, ConvLSTMs have become increasingly popular, achieving state-of-the-art results in various computer vision tasks (e.g.~\cite{CLSTM,Wang2018a,xu2018youtube}). In this work we augment a state-of-the-art architecture for image saliency~\cite{Pan2017} by adding a ConvLSTM module within its internal structure, similar to \cite{Wang2018a,gorji2018going}. More interestingly, we also test a much simpler method for temporal stability. We wrap a convolutional layer with a temporal exponential moving average (EMA)~\cite{EMA} operation. Using this recurrence, the output will always be a smoothed average of its previous states. This method is already used in gradient descent with momentum~\cite{momentumDL} to speed up convergence, replacing the current gradient with the exponential moving average of current and past gradients, derived from mini-batches of the data. To the best of our knowledge, this is the first time that this method has been applied within the architecture of a neural network.  

Ablation studies are commonly used to better understand the performance impact of added components. Whilst this has merit, we propose that simple functions should also be used to investigate the necessity of complex modifications. To this end, in this work we consider both an elaborate ConvLSTM recurrence and a much simpler weighted average recurrence, and show that the simpler approach competes with the ConvLSTM on the task of video saliency.


\section{Related Work}
Video saliency prediction with deep neural networks has basically adapted to this task the architectures proposed for video action recognition.
A first popular option are two-stream networks \cite{simonyan2014two}, in which the motion information is encoded by a pre-computation of the optical flow and adding it in a separate tower from the RGB channels. This is the approach adopted by STSConvNet \cite{bak2017spatio}.
This solution presents two important limitations: the computation overhead that is necessary to compute the optical flow, and the lack of temporal perspective further than the pairs of consecutive frames typically considered when computing optical flow. These shortcomings are partially addressed with the neural architectures where the temporal relation across frames is computed by a recurrent neural network (RNN) \cite{donahue2015long}.
RNN-based deep models for saliency prediction have already been explored \cite{bazzani2016recurrent,jiang2017predicting,Wang2018a,gorji2018going} and are the core of the state of the art solutions.
Similarly to \cite{montes2016temporal} for activity detection, RMDN \cite{bazzani2016recurrent} combined the short-term memory encoded by C3D spatio-temporal convolutions \cite{tran2015learning} with a long short-term memory encoded by a plain LSTM.
However, most current works have adopted a ConvLSTM layer as temporal recurrence, so that the recurrent layer would have a notion of space at a local scale.
The OM-CNN model proposed in \cite{jiang2017predicting} fuses the RGB and optical flow from two-stream architecture with two ConvLSTMs.
The authors of the largest dataset for video saliency prediction, the DHF1K (Dynamic Human Fixation 1K) dataset\cite{Wang2018a}, trained a deep neural model based on ConvLSTM layers with attention (ACLNet). 
The authors of \cite{gorji2018going} exploit an existing model pre-trained for static saliency prediction, but with a more complex architecture composed of four branches fused with a ConvLSTM.

Our model outperforms the presented state of the art with a simple architecture that only considers RGB frames as input.
As in some of the referred works, we exploit a model pre-trained with static images and study its enhancement with two types a temporal recurrence. 


\section{Architecture}

The adopted neural architecture follows an encoder-decoder scheme that processes the temporal recurrence in the bottleneck. The topology of both encoder-decoder is adopted from SalGAN~\cite{Pan2017}, the current top performing static saliency model on the DHF1K saliency benchmark. SalGAN encoder corresponded to the popular VGG-16 convolutional network~\cite{VGG} designed and trained to solve an image classification task. At the decoder side, SalGAN used the same layers as in the encoder in reverse order, and interspersed by upsampling instead of pooling operations. The original SalGAN model was trained using a combination of adversarial and binary cross entropy (BCE) loss. Here, for simplicity, we use only BCE and term the resulting architecture \textit{SalBCE}.


\begin{figure}[t]
\begin{center}
\includegraphics[width=0.65\textwidth]{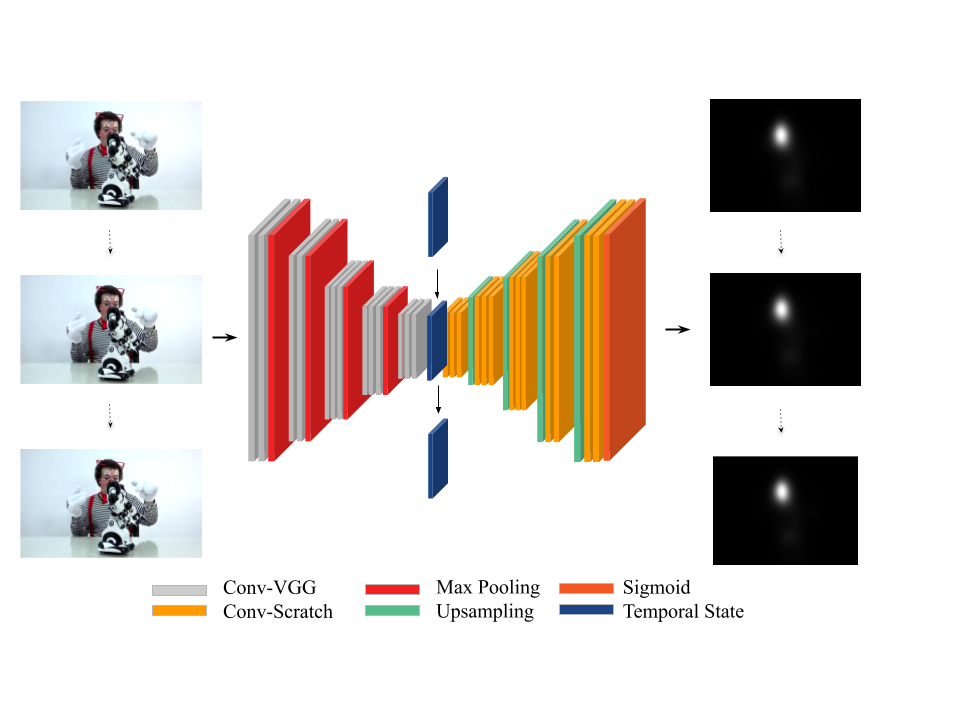}
\caption{Architecture of the our model. A frame is input to the model at each time step. Information encoded from the past frames persists via our recurrence that is located deeper in the network. The output is a per-frame saliency map.}
\label{fig:model_arch}
\end{center}
\end{figure}

We introduce a temporally aware component into the SalBCE network. This is either the addition of a ConvLSTM layer or an exponential moving average (EMA)  applied on a pre-existing convolutional layer. Figure~\ref{fig:model_arch} presents a schematic of our architecture.

\subsection{ConvLSTM}  
\begin{figure}[t]
\begin{center}
\includegraphics[width=0.4\textwidth]{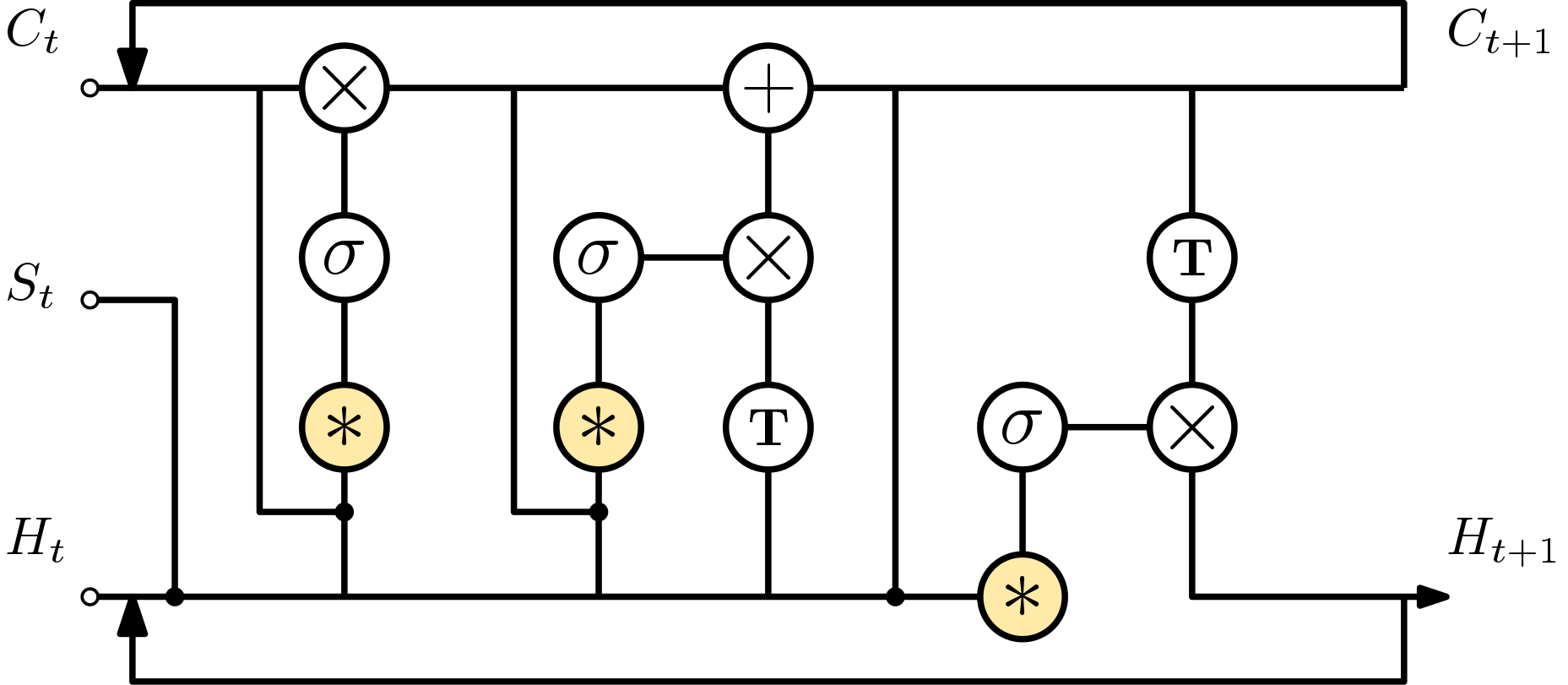}
\includegraphics[width=0.4\textwidth]{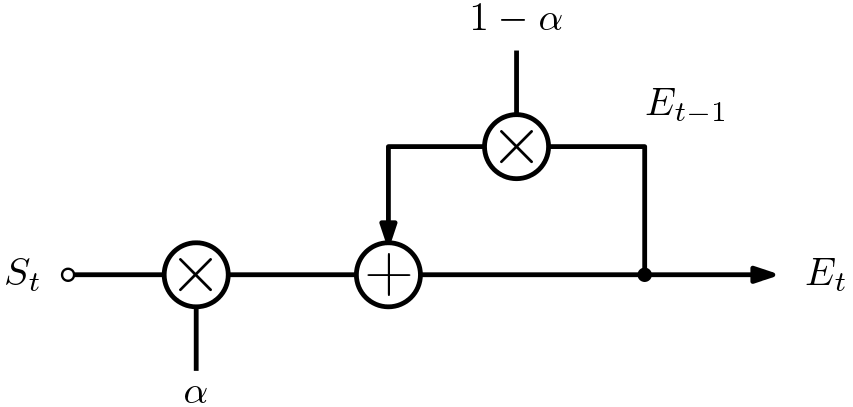}
\end{center}
\caption{\textit{(Left)} LSTM recurrence. Parametric operations are highlighted in yellow. \textit{(Right)} EMA recurrence}
\label{figEMA_LSTM}
\end{figure}

An LSTM is an autoregressive architecture that controls the flow of information in the network using 3 gates: update, forget, and output (Figure~\ref{figEMA_LSTM},~\textit{left}). In ConvLSTMs~\cite{CLSTM}, the operations at each gate are convolutions. Temporal information is preserved in the cell state \textit{C\textsubscript{t}} upon which gated element-wise operations are performed by the update and forget gate. The hidden state \textit{H\textsubscript{t}} is concatenated with the input at each step and propagated through linear and non-linear operations at the gates. At each gate the current state \textit{S\textsubscript{t}} of the model is passed through the ConvLSTM gates and the cell state \textit{C\textsubscript{t}} and hidden state \textit{H\textsubscript{t}} are updated. In the following equations `$\circ$' represents the element-wise product, `$\ast$' a convolution operation, `$\sigma$' the sigmoid logistic function and `$\tanh$' the hyperbolic tangent. 
The \textbf{update}, \textbf{forget}, and \textbf{output} gates can be written as:
\begin{align}
u_t &= \sigma(W_u^S\ast S_t + W_u^H\ast H_{t-1}+W_u^C\circ C_{t-1} + b_u) \\
f_t &= \sigma(W_f^S \ast S_t + W_f^H \ast H_{t-1} + W_f^C \circ C_{t-1} + b_f) \\
o_t &= \sigma(W_o^S \ast S_t + W_o^H \ast H_{t-1} + W_o^C \circ C_{t-1} + b_o)
\end{align}
and the new cell state $C_t$ and hidden state $H_t$ are then given by:
\begin{align}
C_t &= f_t \circ C_{t-1} + u_t \circ \tanh(W_C^S \ast S_t + W_C^H \ast H_{t-1} + b_C) \\
H_t &= o_t \circ \tanh(C_t)
\end{align}
where $W_*^*$ and $b_*$ are the model parameters. 
%

We added the ConvLSTM architecture at the bottleneck of our model, so that the input to the ConvLSTM is an encoded representation of the frame at time $t$. The output cell state $C_t$ is fed to the decoder for further processing that results in a saliency map. To obtain the saliency map, a $1\times1$ convolution is used at the final layer of the decoder, so as to filter out all channels but one. We sequentially pass video frames to the model as input and get a sequence of time-correlated saliency maps in the output. The ConvLSTM component learns to leverage the temporal features during training. The name we gave to this type of model is \textit{SalCLSTM}.

\subsection{Exponential Moving Average} 
As an alternative approach, the exponential moving average (EMA) recurrence \cite{EMA} is added on a specified layer so that at time $t$ the convolutional state of this layer will be a decaying weighted average of the current and all previous states (Figure~\ref{figEMA_LSTM},~\textit{right}). At time $t$ the convolutional layer $S_t$ outputs a state that is fed to the exponential weighted average. The output $E_t$ is then propagated further in the model. Note that there is a hyperparameter $\alpha$ that affects the impact of previous states on the current time step (the lower the value the higher the impact).
\begin{equation}
E_t = \alpha S_t + (1-\alpha)E_{t-1}
\end{equation}

This recurrence is straightforward to implement, especially compared to the ConvLSTM. We experimented with the placement of the EMA function at several different layers with $\alpha=0.1$. We name our model \textit{SalEMA}. On the initial step, where there is no past information, the model runs like a static saliency map predictor. 


\section{Training}

The parameters of SalCLSTM and SalEMA were estimated by backpropagating a pixel-wise content loss that compared the value of each pixel in the predicted saliency map with its corresponding pixel in the ground truth map. The total binary cross entropy loss was computed as the average of the individual binary cross entropies (BCE) over all pixels:
\begin{equation}
L\textsubscript{\it BCE} = -\frac{1}{N}\sum_{\it n=1}^{N}P\textsubscript{\it n} \log(Q\textsubscript{\it n})+(1-P\textsubscript{\it n}) \log(1-Q\textsubscript{\it n})
\end{equation}
where \textit{P} represents the predicted saliency map and {\it Q} the ground truth saliency map.

SalCLSTM and SalEMA were not trained from scratch though, as the parameters of the encoder-decoder convolutional layers were adopted from SalBCE.
SalBCE was trained for 27 epochs over the SALICON~\cite{SALICON} dataset of still images using only the same BCE loss. We also utilized data augmentation techniques (mirroring and rotation of frames) which resulted in improved performance.

Our next step was adding recurrence that uses the intrinsic temporal information of video datasets and train it with the DHF1K dataset~\cite{Wang2018a}. 
The DHF1K dataset \cite{Wang2018a} contains 700 annotated videos at 640$\times$360 resolution. We extracted frames at their original 30 fps rate, and resized them to 192$\times$256 resolution. We loaded them using a batch size of 10 frames from a single video at a time. By backpropagating the loss through time up to a maximum of 10 frames, we avoid exceeding memory capacity and potential vanishing or exploding gradients. We found it was necessary to initialize the ConvLSTM recurrence with the Xavier initialization method~\cite{Xavier_Initialization}, otherwise this model would converge to black images rather than saliency maps. This was likely due to oversaturation of the sigmoid activation layer. We trained all our models for 7 epochs, where we observed the loss reaching a plateau on our baseline. We used the Adam optimizer \cite{Adam} with a learning rate of $10\textsuperscript{-7}$.

\section{Evaluation}

\begin{table}[t]

\begin{center}
\begin{tabularx}{\textwidth}{Xclllll}
\toprule
 & tuned on DHF1K & AUC-J  &s-AUC & NSS 	&  CC  & SIM\\
\midrule
SalBCE (Baseline) &\xmark & 0.874 & 0.724 &	2.047 & 0.382 & 0.268 \\
SalBCE &\cmark & 0.880 & 0.632 &	2.285 & 0.420 & 0.339 \\
SalEMA  &\xmark & 0.883 & \textbf{0.734} & 2.144 & 0.400 & 0.276\\
SalEMA &\cmark & 0.883 & 0.685 & \textbf{2.402 }& \textbf{0.435 }& \textbf{0.349}\\
SalCLSTM &\cmark&\textbf{ 0.887} & 0.693 & 2.364 & \textbf{0.435 }& 0.322  \\

\bottomrule
\end{tabularx}
\end{center}

\caption{Performance results on the DHF1K validation set.}
\label{tab:SalEMAvsSalCLSTM}
\end{table}

The effect of temporal recurrences proposed for SalEMA and SalCLSTM was assessed with five different visual saliency metrics: Normalized Scanpath Saliency (NSS), Similarity Metric (SIM), Linear Correlation Coefficient (CC), AUC-Judd (AUC-J), and shuffled AUC (s-AUC). In all cases, a higher value corresponds to a better performance. The reader is referred to~\cite{bylinskii2019different} for a detailed description of these metrics. The reported figures correspond to an average per video, that is, we first compute the metric on each frame, then average across all frames of each video, and we finally average across all videos.

We train and evaluate our models on three video saliency datasets, namely DHF1K~\cite{Wang2018a}, Hollywood-2 and UCF-sports~\cite{Hollywood_UCF}. DHF1K is a large scale dataset with a high diversity of contents and variable length (from 400 frames to 1200 frames at 30fps). It includes 1000 videos, out of which 700 are publicly annotated, and 300 are withheld for testing purposes. In contrast to DHF1K, Hollywood-2~\cite{Hollywood-Origins} and UCF-sports \cite{soomro2014action} are limited to human actions and can be categorized as task-driven, given that the observers were explicitly asked to identify actions and scene context. These datasets were originally formed for the task of action recognition and were later adopted as a video saliency benchmark. Furthermore, both datasets have been divided into separate shots, so that no scene change occurs in the sequences that are fed into the models. Hollywood-2 is split into a training set of 3100 clips and a test set of 3559 clips, while UCF-sports has been split to a training set of 104 clips and a test set of 48 clips. These shots are much smaller in size than a DHF1K video sample, ranging from 40 frames to just a single frame per shot. We also use SALICON~\cite{SALICON}, a large-scale image saliency database, to set a baseline.
DHF1K is used for experimenting with variations over the proposed models, as well as for comparison with the state of the art together with Hollywood-2 and UCF-sports.


The results in Table \ref{tab:SalEMAvsSalCLSTM} indicate that the simple addition of EMA even without extra training does almost as well as a sophisticated ConvLSTM recurrence, and even improves it after being fine-tuned with the DHF1K training partition. EMA essentially performs a smoothing over the frames of the video by averaging. A possible explanation for why this boosts performance in video saliency is that saliency tends to be relatively consistent across frames, with the exception of rapid movements. 

Encouraged by the positive results of our EMA modification at the bottleneck (layer 30), we explored more possible locations of the EMA function. In particular we tested its placement on: output (layer 61), decoder (layer 56), encoder (layer 7). We also implemented a variation that integrates EMA at two separate layers simultaneously, one in the encoder (7) and one in the decoder (56). In that case we set $\alpha$ to 0.3 at each location so as to not have an over-smoothing effect that would result in a significant lag at adapting to changes in the scene. 
Furthermore, in a video there can be spontaneous scene changes. In such instances, it would be optimal to have the EMA reset and forget all the previous states. However, EMA is not adaptive in this way, so we experimented with a skip connection that allows information to bypass this layer instead~\cite{ResNet}. We also applied a second type of regularization, the dropout technique~\cite{Dropout}, at the convolutional layer right before the EMA layer. Dropout essentially turns off neurons with a preassigned probability (0.5) at each training step. This mitigates co-adaptation of neurons during training, allowing for clusters of neurons to learn independently. This way, at test time, we get the average from an ensemble of layers at location 30. The average of this ensemble pertains to spatial information, but since we are also using EMA, we get the moving average across the temporal dimension as well. 
The results reported in Table~\ref{tab:SalEMA} do not show a clear winning configuration across the five metrics metrics but, as NSS and CC are considered as the most appropriate ones to capture viewing behavior \cite{bylinskii2019different}, we adopted SalEMA30 with dropout as our best configuration. 

\begin{table}[t]
\begin{center}
\begin{tabularx}{\textwidth}{Xclllll}
\toprule
\textbf{Model} & tuned on DHF1K &\textbf{AUC-J} &\textbf{s-AUC} & \textbf{NSS} 	&  \textbf{CC}  & \textbf{SIM}\\
\midrule

SalEMA30 &\xmark & 0.883 & 0.734 & 2.144 & 0.400 & 0.276\\
SalEMA30 &\cmark & 0.883 & 0.685 & 2.402 & 0.435 & 0.349\\
\midrule
SalEMA30 (dropout)&\cmark&	0.886&	0.690&	\textbf{2.495}&	\textbf{0.450}&	\textbf{0.360}\\
SalEMA30 (residual)&\cmark&	0.875	&0.670	&2.274	&0.415	&0.339\\
\midrule
SalEMA61 &\xmark &	0.884&	\textbf{0.737}&	2.133&	0.399&	0.270\\
SalEMA61 &\cmark & \textbf{0.888} & 0.681 & 2.394 & 0.438 & 0.354\\
SalEMA54 &\xmark &	0.883&	0.734&	2.149&	0.401&	0.276\\
SalEMA7 &\xmark	&0.872	&0.656	&2.217	&0.409	&0.318\\
SalEMA7\&54 &\cmark &	0.828&	0.561&	1.403&	0.366&	0.344\\

\bottomrule
\end{tabularx}
\end{center}

\caption{Performance of SalEMA variants on DHF1K.}
\label{tab:SalEMA}
\end{table}

Furthermore, we evaluated our two models on Hollywood-2 and UCF-sports ~\cite{Hollywood_UCF}. 
We compare our models to the current state-of-the-art as evaluated on the test split of the corresponding datasets by Wang \etal~\cite{Wang2018a}. 
Like ACLNet~\cite{Wang2018a}, our models were trained first for DHF1K in all cases, and later fine-tuned for the specific Hollywood-2 or UCF-Sports dataset. 
Table \ref{tab:SoA} shows how, for DHF1K, SalEMA achieves the best performance compared to other models in the current benchmark across all metrics but s-AUC. On the other hand, SalCLSTM obtains the best results on all metrics for UCF-Sports and leads the performance on AUC-J, NSS and CC for Hollywood-2.

\begin{table}[t]
\begin{center}
\begin{tabularx}{\textwidth}{lXlllll}
\toprule
\textbf{Dataset}&\textbf{Model}  & \textbf{AUC-J}  &\textbf{s-AUC} & \textbf{NSS} 	&  \textbf{CC}  & \textbf{SIM}\\
\midrule
\multirow{5}{*}{DHF1K}&SalEMA & \textbf{0.890}	& 0.667 & \textbf{2.573} & \textbf{0.449} & \textbf{0.465}\\
&SalCLSTM & 0.887 & 0.693 & 2.364 & 0.435 & 0.322  \\
&ACLnet \cite{Wang2018a}& \textbf{0.890}	&	0.601	&	2.354		&	0.434 &	0.315	 \\
&SalGAN \cite{Pan2017} & 0.866	& \textbf{0.709}	& 2.043	& 0.370	 & 0.262\\
&DVA \cite{DVA}& 0.860 & 0.595	& 2.013	& 0.358		& 0.262\\
\midrule
\multirow{5}{*}{Hollywood-2}&SalEMA & 0.919	&	0.708	&	3.186	&	0.613 &	0.487\\
&SalCLSTM  & \textbf{0.933} & 0.715	& \textbf{3.499}	& \textbf{0.672}	& 0.530 \\
&ACLnet \cite{Wang2018a} & 0.913	& \textbf{0.757} & 3.086 & 0.623 & \textbf{0.542} \\
&OM-CNN \cite{jiang2017predicting} & 0.887	& 0.693	& 2.313	& 0.446	 & 0.356\\
&DVA \cite{DVA}& 0.860 & 0.727	& 2.459	& 0.482		& 0.372\\
\midrule
\multirow{5}{*}{UCF-sports}&SalEMA & 0.906	&	0.740	&	2.638	& 0.544 &	0.431\\
&SalCLSTM &\textbf{0.914}	&\textbf{0.782} & \textbf{3.063} & \textbf{0.611} & \textbf{0.477} \\
&ACLnet \cite{Wang2018a} &0.897	& 0.744	& 2.567&	0.51&	0.406 \\
&DVA \cite{DVA}& 0.872 & 0.725	& 2.311	& 0.439		& 0.339\\
&OM-CNN \cite{jiang2017predicting} & 0.870	& 0.691	& 2.089	& 0.405	 & 0.321\\
\bottomrule
\end{tabularx}
\end{center}

\caption{Comparison of SalEMA and SalCLSTM with the state of the art on DHF1K, Hollywood-2, and UCF-sports test sets.}
\label{tab:SoA}
\end{table}

\begin{figure}[t]
\begin{center}
\includegraphics[width=0.45\textwidth]{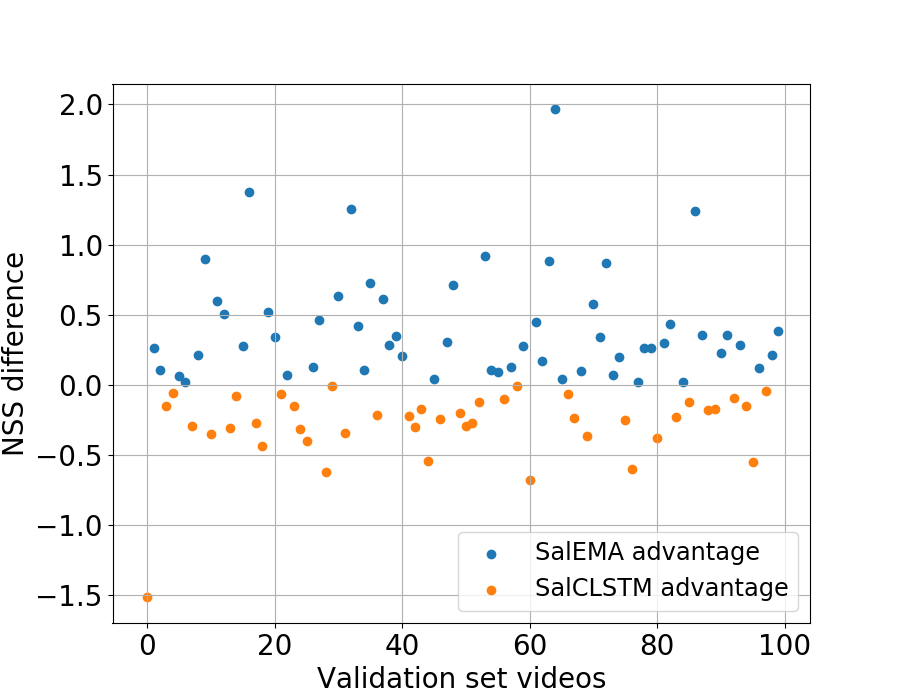}
\includegraphics[width=0.45\textwidth]{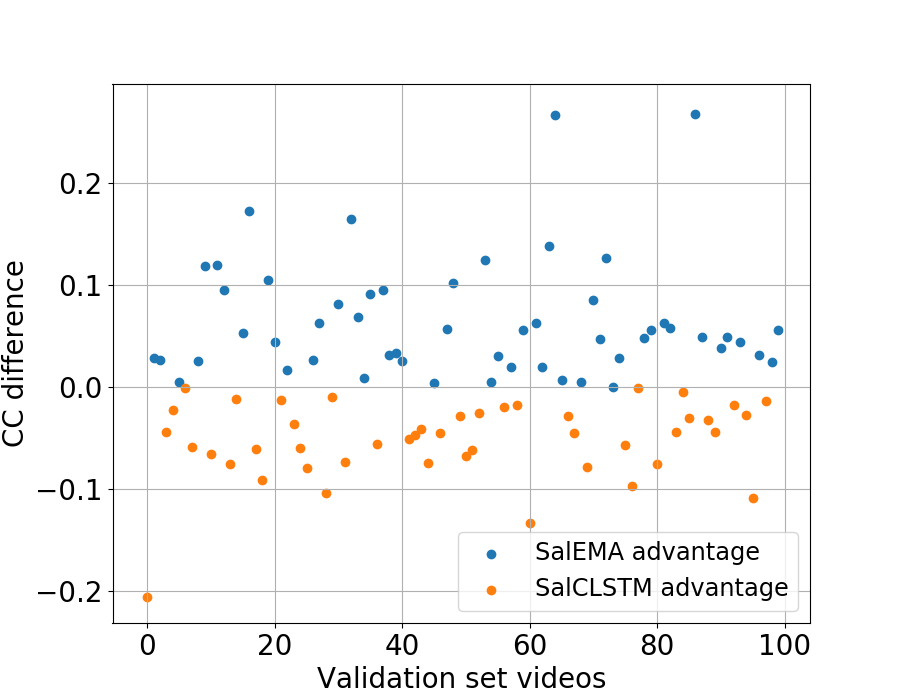}
\end{center}
\caption{Per-video comparison between SalEMA and SalCLSTM using the NSS and CC metric on the DHF1K validation set. The values represent the margin by which a model's performance differs from the other. } 
\label{NSS-comparison}
\end{figure}

A more detailed analysis between SalEMA and SalCLSTM was obtained by plotting the difference in their NSS and CC performance per video in the DHF1K validation set (100 videos). Concretely, we subtracted the metric value achieved by the SalCLSTM from that of SalEMA in each video and display the results in Figure~\ref{NSS-comparison}. This way, we can assess whether the two configurations end up producing similar results.
In this case we would expect the variance to be low and the NSS difference to be close to zero most of the time. However, the results are sparse and diverge from video to video. This observation serves as evidence that the function approximated by the ConvLSTM is differs from that of an exponential moving average, despite its similar overall effectiveness.

\begin{figure}[t]
\begin{center}
\includegraphics[width=\textwidth]{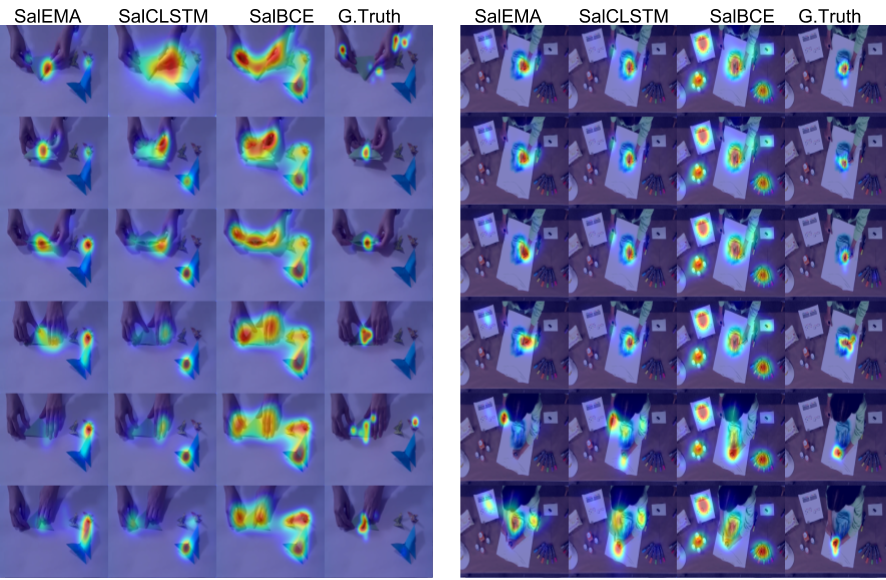}

\end{center}
\caption{We picked two samples that showed high divergence in performance between the two methods and visualized the predictions. SalEMA did much better than SalCLSTM on the sample displayed on the left side, while it was the opposite for the other sample. The images displayed correspond to intervals of 100 frames in the video.} 
\label{QResults}
\end{figure}

\begin{table}[t]
\begin{center}
\begin{tabularx}{\textwidth}{XXXXXl}
\toprule
\textbf{\textbf{$\alpha$}} & \textbf{AUC-J}  &\textbf{s-AUC} & \textbf{NSS} 	&  \textbf{CC}  & \textbf{SIM}\\
\midrule

\textit{0.05} & \textbf{0.886}	& 0.687	& 2.470	& 0.448	& 0.358\\
\textit{0.1} & \textbf{0.886}	& \textbf{0.690}	&	\textbf{2.495}	&	\textbf{0.450}	&	\textbf{0.360}\\
\textit{0.2} & 0.885	&	0.688	&	2.476	&	0.446	&	0.358\\
\textit{0.3} & 0.884	&	0.685	&	2.451	&	0.442	&	0.356\\

\bottomrule
\end{tabularx}
\end{center}
\caption{Sensitivity of SalEMA30 to $\alpha$ on DHF1K validation.} 
\label{tabAlpha}
\end{table}

\begin{figure}[t]
\begin{center}
\includegraphics[width=\textwidth]{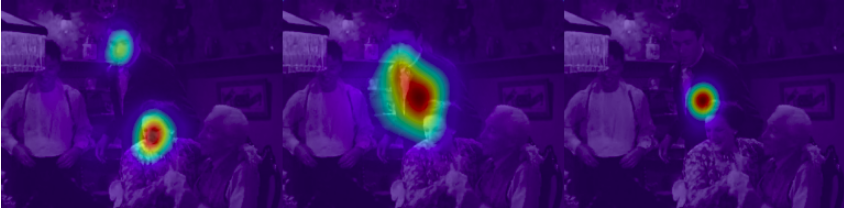}
\includegraphics[width=\textwidth]{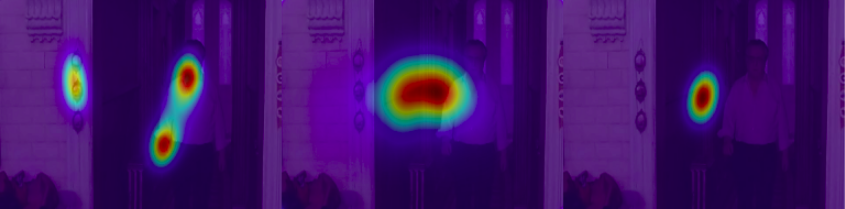}
\end{center}
\caption{Predictions from two Hollywood outliers where SalEMA performed particularly bad. The order corresponds to: SalEMA (left image), SalCLSTM (middle image), ground truth (right image). The ground truth appears aberrant, as it completely ignores human faces that are well-known to be salient objects.} 
\label{figHollywoodQ}
\end{figure}

We also delved deeper into the Hollywood-2 dataset for potential clues that would explain the difference in performance. This dataset consists of very small shots, including even single-frame shots. In these cases we found that the ConvLSTM does much better than the EMA (by a margin of around $4$ NSS points). We also noticed, however, that in these cases the ground truths for the saliency maps correspond to a central Gaussian, despite the fact that other salient objects are present in other locations of the frame.
Figure~\ref{figHollywoodQ} shows two examples in which the provided ground truth focuses in the center, although different faces appear in the image. In these examples, SalEMA captures these salient objects better, while SalCLSTM seems to focus on the center.

Finally, we experimented with the $\alpha$ hyperparameter by varying its value and also by making it trainable. Table~\ref{tabAlpha} shows relatively stable performance despite the variations on the value. We also had our model learn alpha on its own by introducing a trainable parameter $p$. To ensure that the resulting update equation represents a convex combination of the current features and previous state, $p$  is passed through a sigmoid so that the final value is constrained to $[0,1]$. The resulting recurrence is:
\begin{equation}
E_t = \sigma(p) S_t + (1-\sigma(p))E_{t-1} 
\end{equation}

Whereas all other parameters of the model are set to a learning rate of $10\textsuperscript{-7}$, the learning rate of alpha was set to 0.1 and was trained separately for 3 epochs on SalEMA pretrained with $\alpha=0.1$. We set $\sigma(p)$ to 0.5 at the start of this tuning and by the end, it converges to 0.1477. The final performance was found to be approximately the same as the best model in Table~\ref{tabAlpha}.


\section{Conclusions}

This work has presented SalEMA and SalCLSTM, two variations of a convolutional neural network for video saliency prediction. 
Their main difference is how temporal recurrence is modelled, whether with a simple yet effective exponential moving average with a single parameter, or a convolutional LSTM that despite being adopted for many video sequence processing tasks, seems needlessly complex for this specific task of video saliency prediction. This indicates that, in some cases, components of more sophisticated models may just learn to approximate much simpler functions. It is likely that similar methods can be conceived of in other types of tasks as well.


On another note, ablation studies are a common practice for evaluating the contribution that an added component has on a model's performance. We argue that there should be a more detailed effort in analyzing the behavior of deep architectures. Using predefined functions like the one presented in this work may shed more light on the necessity of a complex architecture.

\paragraph{Acknowledgements}
This publication has emanated from research conducted with the financial support of Science Foundation Ireland (SFI) under grant number SFI/15/SIRG/3283 and SFI/12/RC/2289. This work has been developed in the framework of project TEC2016-75976-R, funded by the Spanish Ministerio de Economia y Competitividad and the European Regional Development Fund (ERDF).

\bibliography{bmvc_final}
\end{document}